\documentclass[letterpaper, 10 pt, conference]{ieeeconf}  

\IEEEoverridecommandlockouts                              

\usepackage{graphics} 
\usepackage{graphicx}
\usepackage{subcaption}
\usepackage{amsmath} 
\usepackage{amssymb}  
\IEEEtriggeratref{6}

\title{\LARGE \bf
Self-Evaluation in One-Shot Learning from Demonstration of Contact-Intensive Task
}

\author{Mythra V. Balakuntala$^{1}$, L. N. Vishnunandan Venkatesh$^{1}$, Jyothsna Padmakumar Bindu$^{1}$, Richard M. Voyles$^{2}$
\thanks{*This work was supported by NSF Center on RObots and SEnsors for HUman well-Being (RoSe-HuB) and Army ForWard grant.}
\thanks{$^{1}$Mythra V. Balakuntala, L. N. Vishnunandan Venkatesh and Jyothsna Padmakumar Bindu are with School of Engineering Technology, 
        Purdue University, IN 47907, USA
        {\tt\small mbalakun@purdue.edu, lvenkaten@purdue.edu, jpadmaku@purdue.edu }}%
\thanks{$^{2}$Richard M. Voyles is Professor of Robotics with the School of Engineering Technology, Purdue University, IN 47907, USA
        {\tt\small rvoyles@purdue.edu}}%
}
\begin{document}

\maketitle
\thispagestyle{empty}
\pagestyle{empty}
\begin{abstract}
Humans naturally ``program`` a fellow collaborator to perform a task by demonstrating the task few times. It is intuitive, therefore, for a human to program a collaborative robot by demonstration and many paradigms use a single demonstration of the task. This is a form of one-shot learning in which a single training example, plus some context of the task, is used to infer a model of the task for subsequent execution and later refinement. This paper presents a one-shot learning from demonstration framework to learn contact-intensive tasks using only visual perception of the demonstrated task. The robot learns a policy for performing the tasks in terms of a priori skills and further uses self-evaluation based on visual and tactile perception of the skill performance to learn the force correspondences for the skills. The self-evaluation is performed based on goal states detected in the demonstration with the help of task context and the skill parameters are tuned using reinforcement learning. This approach enables the robot to learn force correspondences which cannot be inferred from a visual demonstration of the task. The effectiveness of this approach is evaluated using a vegetable peeling task.
\end{abstract}

\section{INTRODUCTION}
For collaborative robots to become ubiquitous in households and industries, there is a greater need to reduce their explicit programming to perform varied tasks, by taking advantage of the different social learning strategies \cite{cakmak2010exploiting}. Learning from demonstration is one such approach, where task demonstration is carried out in a way that is very intuitive to humans and hence can be done with little to no training. LfD has progressed from a pure record and playback approach in the past to a learning based one which can achieve generalization. We use LfD as a goal-based imitation technique wherein a task-expert demonstrates the task to the robot from which the  robot identifies the task intent and make inferences about the sub-tasks, the sequence of sub-tasks and the states of the object under manipulation. The robot then performs the entire task through this learned sequence of sub-tasks by checking for desired state changes at the end of each sub-task rather than simply mimicking the set of manipulations displayed during the demonstration. This breakdown into sub-tasks is not programmed into the robot but the robot learns to identify the task as a composition of pre-learnt tasks or a priori skills. Thus, LfD achieves generalization by offering an efficient and flexible framework to easily and quickly extend the capabilities of collaborative robots to perform new and complex tasks.\\
\begin{figure}
    \centering
    \includegraphics[width=0.48\textwidth]{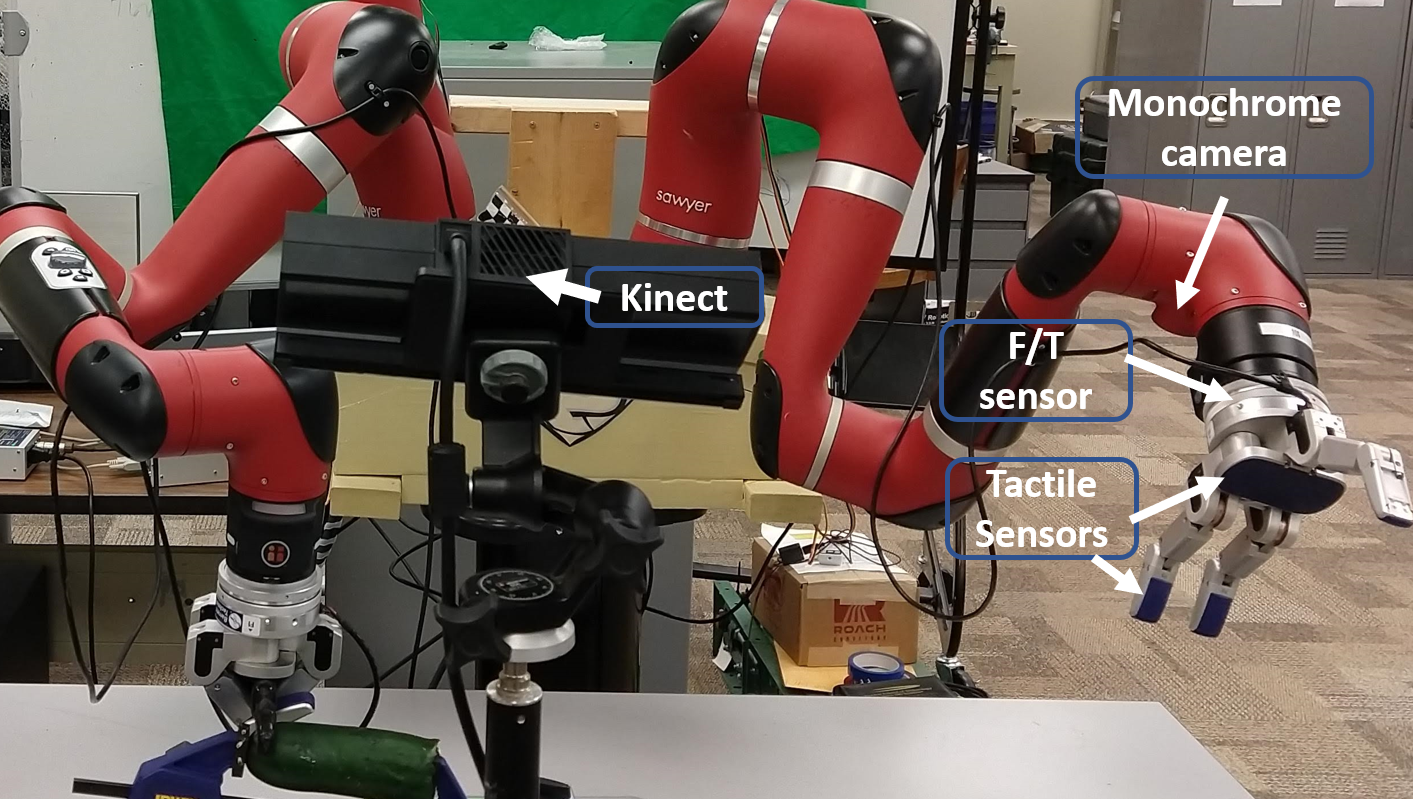}
    \caption{The robot testbed with multimodal sensing capabilities}
    \label{fig:rob}
\end{figure}
\begin{figure*}[t]
    \centering
    \includegraphics[width=0.7\textwidth]{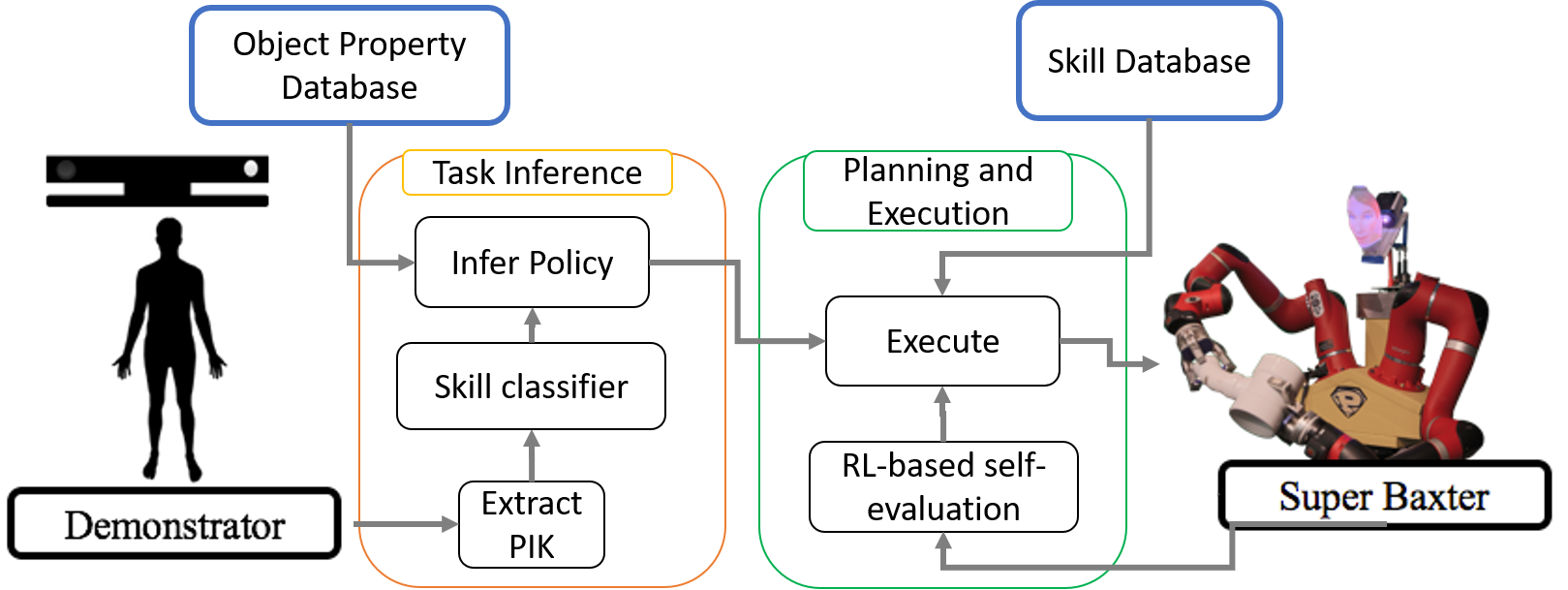}
    \caption{Architecture of the proposed one shot learning framework}
    \label{fig:arch}
\end{figure*}
One of the main challenges for collaborative robots in domestic or industrial settings is the inherent difficulty to perform contact-intensive tasks. Our work presents a novel method to make a robot perform a contact-intensive task by using LfD with self-evaluation from just one visual demonstration. It is challenging for the robot to infer forces involved in such tasks from visual perception alone. Demonstrations are performed by task-experts and not domain-experts in robotics. For task-exports visual demonstrations are more intuitive than kinesthetic demonstrations. Kinesthetic teaching involves the task-expert physically guiding the robot manipulator to make it perform the task. Learning from a single demo further increases the ambiguity in the task inferences that can be made. Despite these limitations, our LfD approach that utilizes extraction of interaction key points was successful in achieving skill goals as well as the overall task goal in a contact-intensive task. Towards the purpose of generalizing the task to multiple objects, the proposed method also implements a self-evaluation step at the end of each skill or sub-task which attempts to tune the individual skill parameters using a reinforcement learning approach. This evaluation is currently based on skill completion rather than optimal skill performance. Hence the evaluation metric used in the present work is binary, i.e. whether the skill was completed or not. The reinforcement learning algorithm adopts a "greedy-towards-skill-goal" approach which focuses on the immediate completion of skill rather than the exhaustive exploration of action set. Further, LfD paradigms generally use the same environment for both human and robot, but another novelty of this work is that the demonstration of the task can be performed in a simulated and inert environment.
\section{RELATED WORK}
Early works on LfD involved record and playback methods \cite{dufay1984approach}. The demonstrated task was decomposed into a  sequence  of  state  transitions which was identified to be achieved through a series of actions from a given set. And these state-action correspondences were programmed as  if-then  rules. Later works used machine learning \cite{muench1994robot} and neural networks \cite{argall2011teacher}\cite{derimis2002imitations} in LfD for inference. Other approaches to learning tasks involve using demonstrations to learn the rewards and the states and use reinforcement learning to learn a policy to achieve the desired states. These methods are called inverse reinforcement learning or inverse optimal control \cite{abbeel2004apprenticeship}\cite{billard2016learning} where rewards to achieve the task are learnt from the demonstrations. Traditionally, learning from demonstration has been mostly used for kinematic tasks like pick and place, where reinforcement learning has been used to achieve completion as well as optimal performance of the motion primitives \cite{stulp2012reinforcement}.\\

Another important aspect in LfD approaches is the inteface used for teaching. Multiple demonstration interfaces like sensorized gloves \cite{voyles1999gesture} have also been attempted in LfD with most of the approaches in literature using kinesthetic (hand-holding) or teleoperated teaching of robot. Kinesthetic teaching eliminates the problem of correspondence, which is the mapping of demonstrator motions to robot with a different physical structure, but it is not a very intuitive method for task experts with no knowledge of the robot but has rich task knowledge. Hence, vision based approaches are preferred over kinesthetic/teleoperation methods for teaching. Vision also helps the LfD paradigm progress from purposeless imitation to inference based generalized task performance. Using vision based approaches results in additional complexity of not being to estimate forces/tactile information from demonstrations for contact-intensive tasks \cite{elliott2017learning}. One solution to this is to provide prior knowledge of environment, i.e. states and skills \cite{levine2015learning}. Another technique is to extract geometric constraint based interaction phases to represent relation between objects in the scene \cite{baisero2015temporal}. Here we propose a more generalized approach to learn a gross policy from demonstration based on physical interactions among agent and objects, and then self-learn the corresponding forces needed using reinforcement learning.\\

The results from our self-skill-evaluation experiments indicate that introducing coaching in our LfD approach shall improve task performance. Human-in-the-loop evaluation as in \cite{elliott2017efficient}\cite{mollard2015robot} can help us obtain more accurate skill parameters as well as better transition between skills. Such coaching can be done through speech, gestures or partial demonstrations thus eliminating the need for multiple  demonstrations.

\section{Task Learning from demonstration}
In this section, we describe the learning from demonstration paradigm proposed in this paper. The learning begins with one visual demonstration of the task. A task policy is learnt from this single demonstration as a sequence of sensorimotor primitives or skills that the robot knows to perform. This policy is then executed by the robot followed by self-evaluation of the skills. Based on skill performance, the skill parameters are tuned to achieve desired goal state.

\subsection{Inference from vision}
The demonstration obtained is in the form of a single RGB-D video recorded using Microsoft KINECT. The robot has a pre-learnt database of objects. This dataset contains features and possible states for different objects. The features include shape, mass, stiffness, object detection features and states like peeled or unpeeled for vegetables, relative position, orientation and  filled/empty status of containers. The vision system has the ability to detect these objects and their states from the demonstration.\\

We draw inspiration from one shot gesture recognition paradigms which use key points in the kinematic trajectories like inflection points \cite{cabrera2017one} or mixed features around sparse keypoints \cite{wan2016explore} in RGB-D videos to classify the gestures in one-shot. We propose a key point for identifying physical interactions between objects and the agent in the scene. This is based on identifying the contact condition between an object closest to the hand and the hand, using the hand region of interest, object, wrist and hand tip positions.\\

\subsubsection{Physical Interaction Keypoints (PIK)} The policy inferred is based on trajectories of agent's hand and object states observed in the demonstration. The hand position is extracted from the agent's skeleton obtained from the RGB-D video using the pyKinect library. Then in each frame, a region of interest sphere is constructed around the agent's hand. Let the number of objects in the scene be $n$. The workspace $W$ is partitioned into $n$ voronoi spaces $V_i$ with object centroids as seeds, so $W = V_1 \cup V_2 \cup \dots V_n $. Then the contact condition $\psi$ between the hand and the object whose voronoi space the hand lies in is computed using the wrist and hand tip position. This is a binary feature which indicates either a contact (1) or no contact (0). A similar feature $\phi$ is computed for only those objects in region of interest (ROI) of the hand as well. For each object in the ROI, the contact condition $\phi$ with nearest object (excluding hand) is computed. This results in two contact features. The frames where the transition of these feature values  occur i.e. either $\phi_i$ or $\psi_i$ flips, is designated as  physical interaction keypoints. These points indicate make or break of contact between object and agent or between objects.\\

The PIKs indicate the change in skills and are used for temporal segmentation of the entire demonstration into multiple segments $\Theta_i$, with each PIK representing the point of segmentation. We construct features based on interactions to classify each of these segments into apriori skills. For each of the segments, relative motion trajectories $X_i$ between hand and the object it is interacting with are extracted. If the hand is not interacting with any object i.e. $\kappa = 0$ then the relative motion trajectories are computed with the object it interacts with in segment $\Theta_{i+1}$. Then each segment represented as $$\Theta_i = (\psi_i, X_i, u(\dot{X}_i), u(\dot{Y}_i), \phi_i)$$ where $Y_i$ is the absolute velocity of the hand, and
\begin{eqnarray}
u(\dot{X}_i) = \begin{cases} 1 & \text{if}\quad \dot{X}_i \geq 0\\
0 & \text{otherwise}\end{cases}
\end{eqnarray}
Each object is assigned an ID which is an integer value representing the object. Let the skill class label of segment $\Theta_i$ be $C_i$. Then the set of $\phi_i$, $\psi_i$, $u(\dot{X}_i)$, the class label of previous class $C_{i-1}$, and object ID values is used to classify each segment $\Theta_i$ into an apriori skill using decision trees.\\
The learnt policy is obtained in two steps - Firstly, the sequence of the inferred skill classes is obtained as above. Then this sequence is parsed again to add any skills required for transition from skill $C_i$ to $C_{i+1}$. The final learnt policy is thus a sequence of the apriori skills in the demonstration,
\begin{equation*}\Pi = \{(C_1, s*_1, \Theta_1), (C_2, s*_2, \Theta_2), \dots (C_m, s*_m, \Theta_m)\}\end{equation*}
 where $s*_i$ is the state of object in interaction after skill $C_i$ in the demonstration. This state is the reference goal state for the skill execution. $m$ is the number of skills and $\Theta_i$ the ith demonstration segment. For skills which were added as transition, $\Theta_i$ is based on $C_{i-1}$ and $C_{i+1}$ and the goal state is the state at the beginning of $C_{i+1}$. Each skill $C_i$ is associated with an execution sensorimotor control model to perform the skill. This is described in the following section.
\begin{figure}[t]
    \centering
    \begin{subfigure}[t]{0.1\textwidth}
        \centering
        \includegraphics[width=\textwidth]{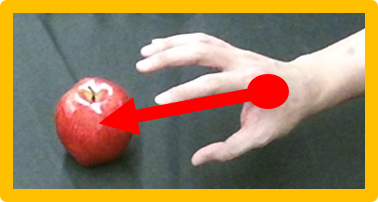}
        \caption{Approach}
    \end{subfigure}
    \begin{subfigure}[t]{0.1\textwidth}
        \centering
        \includegraphics[width=\textwidth]{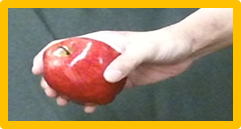}
        \caption{Grasp}
    \end{subfigure}
    \begin{subfigure}[t]{0.1\textwidth}
        \centering
        \includegraphics[width=\textwidth]{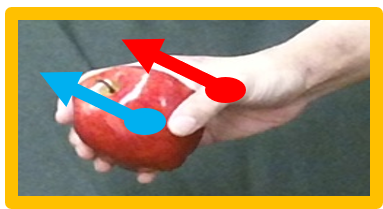}
        \caption{Transport}
    \end{subfigure}
    \begin{subfigure}[t]{0.1\textwidth}
        \centering
        \includegraphics[width=\textwidth]{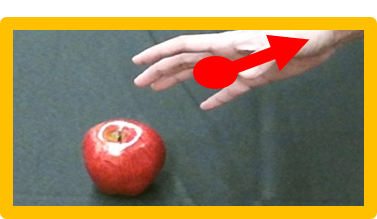}
        \caption{Retract}
    \end{subfigure}
    \begin{subfigure}[t]{0.1\textwidth}
        \centering
        \includegraphics[width=\textwidth]{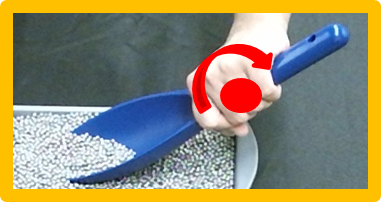}
        \caption{Scoop}
    \end{subfigure}
    \begin{subfigure}[t]{0.1\textwidth}
        \centering
        \includegraphics[width=\textwidth]{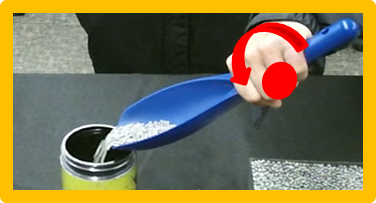}
        \caption{Unscoop}
    \end{subfigure}
    \begin{subfigure}[t]{0.1\textwidth}
        \centering
        \includegraphics[width=\textwidth]{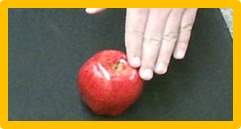}
        \caption{Guarded Move}
    \end{subfigure}
    \begin{subfigure}[t]{0.1\textwidth}
        \centering
        \includegraphics[width=\textwidth]{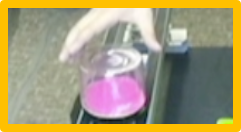}
        \caption{Visual servoing}
    \end{subfigure}
    \begin{subfigure}[t]{0.1\textwidth}
        \centering
        \includegraphics[width=0.8\textwidth]{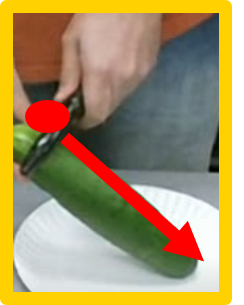}
        \caption{Move with contact}
    \end{subfigure}
    \begin{subfigure}[t]{0.1\textwidth}
        \centering
        \includegraphics[width=0.8\textwidth]{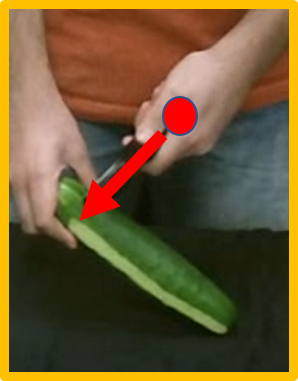}
        \caption{Move to contact}
    \end{subfigure}
    \caption{List of a priori skills.}
    \label{fig:skills}
\end{figure}
\subsection{A priori skills} \label{skills}
A priori skills are pre-learnt atomic sensorimotor control actions which the robot can perform. The learnt policy $\Pi$ is the sequence of skills and associated states at the end of each skill. Fig \ref{fig:skills} shows some of these skills. Each skill is defined  as a control action with a particular sensor feedback and goal condition. The database of skills is segmented into two types - force-based and positional. Force-based skills have an impedance control policy for achieving desired force trajectories and positional skills have a positional control policy as shown in \ref{eqn:ski}.
Let the state of the system be denoted by $s$, the pose at time $t$ by $x_t$, the goal state by $s*$, and the desired pose by $x^d$. Then, the skill control policy for kinematic skills can be defined as follows,
\begin{equation}
    x_{t+1} = x_t + k_1(f(x^d, s*) + k_2 \dot{f}(x^d,s*)) \label{eqn:ski}
\end{equation},
where $k_1$ and $k_2$ are gain parameters, $f$ is the total sensor feedback error function which depends on the desired pose and goal state. The desired pose is defined by the relative position trajectories recorded in $\Theta_i$ and the goal state is the detected state at end of the skill. The feedback error function is a combination of feedback from the different sensing modalities based on the skill being used.The pose vector $x_t$ in our case includes arm pose and hand pose.\\

The impedance-controller is similar to the positional controller except that the controlled variables are joint torques instead of end-pose and gripper state. We use an impedance control policy for force-based skills like \textit{move with contact} as follows, 
\begin{equation}
    \tau = J(\theta_t)^T F_d + K_1(f(x^d, s*) + K_2 \dot{f}(x^d,s*)) \label{eqn:fski}
\end{equation}
where $\tau$ is the 7-dof joint torque vector, $J(\theta_t)$ is the Jacobian at joint configuration $\theta_t$ and $F_d$ is the desired pose at the end-effector. A similar impedance controller with feedback error function $f$ as the error in joint trajectories was used for force-based manipulations of deformable objects in \cite{lee2015learning}. We use a multimodal sensing approach to compute the feedback and detect the state during execution. Different sensing modalities are used based on the skill being executed. For example, the feedback error is a function of tactile forces for \textit{grasp} skill and for \textit{visual servoing}, it is a function of vision based relative position between goal object and agent.

\subsection{Robot Testbed}
The execution of skills was evaluated using Purdue SuperBaxter as the platform \cite{taa2018wm}. The robot is incorporated with multiple sensing modalities which are leveraged to achieve desired self-evaluation. An RGB-D camera (KinectV2), and two monochrome cameras, one on each hand of SuperBaxter, acts as the vision interface of the robot. The hands are Barrett technology BH-282 hands with tactile sensing on the palm and fingers. A 6-dof Force-Torque sensor is integrated in the wrist to estimate the forces felt on the end-effector. These sensors provide multiple modalities of sensing information which are used in combination or isolation based on the skill being executed. User demonstrations are collected through the RGB-D camera of the robot. The robot testbed is shown in Fig. \ref{fig:rob}.  

\subsection{Self-evaluation and Reinforcement learning} 
Our paper deals with the execution of a contact intensive task. To help us execute a contact intensive task through LfD, we adapt to multi modal sensing  to perform self-evaluation and correction of skill parameters. In the context of Learning from Demonstration, vision algorithms are employed to make sense of the demonstration. Tasks dealing with positional inferences are relatively easier to infer from visual perception e.g. Pick and place tasks. In the case of a contact intensive task, we can estimate from the vision system if contact with the object of interest was obtained but we cannot estimate the force acting upon the object with certainty. This necessitates the integration of force sensing to help estimate the force being applied on the object to optimize or tune the skill by means of self-evaluation using reinforcement learning.\\

In essence, the robot performs the learnt policy $\Pi$ following which it uses the vision-based state detection to evaluate if the same goal state $s*$ was achieved. If the goal state is not achieved the system uses reinforcement learning to perform the tuning till it performs the task.\\

Reinforcement Learning deals with an agent and the actions the agent may take. The goal of any reinforcement learning algorithm is finding out the policy (set of actions to take at any given state) that maximizes the agents rewards. This potential reward is a weighted sum of the expected values of the rewards of all future steps starting from the current state. The learning is performed over a finite set of iterations. The states, actions and rewards enlisted below are the input to the learning algorithm. \\
\subsubsection{State Space}
The state space within our reinforcement learning framework is a measure of the environment and object properties which includes various states of an object from the object database. Depending on the environment and the objects present in the environment, the states are obtained. The state of the environment at the end of each skill during the demonstration is taken as the goal state for that skill. This goal state is already present in the object database and is obtained as a result of the objects being detected in the environment by the vision perception system. \\

\subsubsection{Action Space}
From the inference obtained during the demonstration, we obtain a baseline trajectory performed by the user. This baseline trajectory is calibrated to obtain a contact trajectory or an action trajectory, which would achieve a trajectory across the object the robot interacts with at a given contact force. The action space is defined for every skill and  comprises of all these trajectories resulting from obtaining specific contact forces along the trajectory.  The action set however is variable i.e. the robot keeps adding new actions over time as a result of failing to achieve the goal state. A new action is added to the action set when all the existing actions  result in penalties.  The action workspace is inclusive of all these actions present in the action set. \\

\subsubsection{Rewards}
The reward for the reinforcement learner is binary i.e. a low positive reward if desired goal state $s*$ is reached or higher negative reward if it reaches any other state $s' \neq s*$. Let system transition from current state $s$ to new state $s'$ on performing action $a$. Then the reward is given by,
$$ R_a(s,s') = c_1*\delta_{s',s*} - c_2(1-\delta_{s',s*}) $$
where $\delta_{s',s*}$ is the kronecker delta function resulting in $1$ when $s*$ and $s'$ are same and 0 otherwise and $c_2 > c_1>0$\\
\subsubsection{Q Learning}
Q learning is a form of temporal difference-based learning. The goal of q learning is to learn a policy that guides us to perform the best action, given any state. Q Learning uses a table which  holds the q values for every state action pair \cite{Watkins1992}. Each q value is the maximum expected future reward for every state action pair. Q Learning doesn't start off with an initial assumed policy but rather improves upon the policy by updating the q values every iteration with the help of the Bellman equation shown below.The best action at any state is the largest q value corresponding to an action for that state in the q table.\\
\begin{eqnarray}
    Q_{t+1}(s_t,a_t) &=& Q_{t}+\alpha[ r_{t+1}(s_t,a_t) + \gamma~ \max_a Q_t(s_{t+1},a) \nonumber \\&-& Q_{t}(s_t,a_t)]
\end{eqnarray}

In our application of the Q Learning algorithm for self evaluation, we assign a new Q table every time we encounter an object whose properties are completely different from the objects the robot has been dealing with so far. This would ensure that the RL framework can optimize the skill for multiple classes of objects or multiple groups of objects. Thus, each skill can have multiple Q tables depending on the objects the robot interacts with during the skill. Also, keeping in mind that we are dealing with a varying action set, we also adapt with a varying q table when any new action is added to the action set. \\

The Q learning algorithm has three hyperparameters which we need to tune depending on our application. The first is the learning rate. Learning rate determines how much weightage we give to the new q value obtained from performing an action versus the previous q value. A very small learning rate (nearing to zero) would result in no update to the q values as the new updates will be deemed unimportant and vice versa.\\

The second hyperparameter of importance is the discount rate. Discount factor determines how to give higher weight to near rewards received than rewards received further in the future. The reason for using discount factor is to prevent the total reward from going to infinity. The third hyperparameter which deals with exploration and exploitation is mentioned in detail below. \\

\subsubsection{Exploration and Exploitation} \label{exp} 
In our application of self-evaluation, we are 'greedy-towards- skill-goal' completion. Once a new task is shown in demonstration, the robot first generates a q table depending on the states and actions in the environment. The robot starts exploring the environment and if it encounters an action that completes the task, the robot stops the learning process and updates the q table and reports that task is complete. Intuitively at this point, we can say that the last action added to the action set is the one that resulted in task completion and all other actions in the set failed at achieving the task. In other words, the learning process stops as soon as task is completed and this is the first time the robot tries to optimize its skills to learn the task. As the robot performs the task over time it learns and obtains a better understanding about the task and keeps updating the q table, optimizing the skills involved in the task. Thus, we do not focus on exploring the entire action space and updating the q table but rather focus greedily on task completions during the process of exploration and exploitation.\\

Keeping this greedy-towards-skill-goal approach in context we perform exploration and exploitation differently. When the robot tries to perform the task subsequent or multiple times, we first start off  by exploiting the already existing learned policy i.e. performing the optimal contact trajectory action to perform the task. If this action fails in task completion, we go on to performing one of two exploratory procedures. The first procedure  is that we add a new action to our variable action set and update our q table as well to accommodate this new action and perform that action immediately as a form of exploration. The second exploratory procedure is that we choose at random one of our already existing actions and see how it impacts the environment and update our q table accordingly.\\ 

\begin{figure}[t]
    \centering
    \begin{subfigure}[t]{0.5\textwidth}
        \centering
        \includegraphics[width=0.9\textwidth]{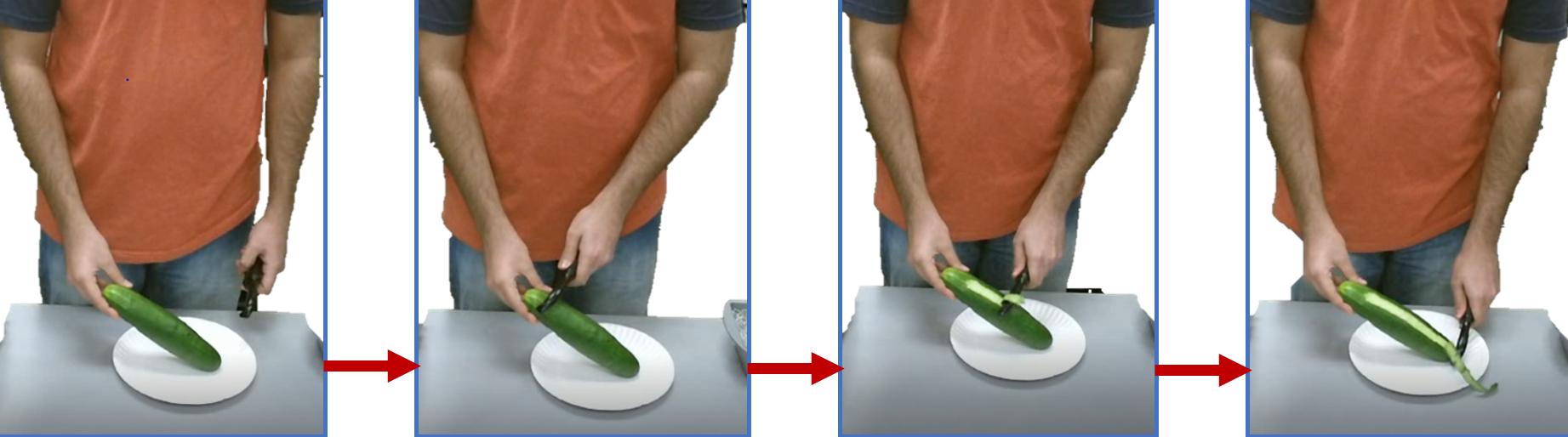}
        \caption{Demonstration of task}
        \label{fig:demo}
    \end{subfigure}\\
    ~ 
    \begin{subfigure}[h]{0.5\textwidth}
        \centering
        \includegraphics[width=0.9\textwidth]{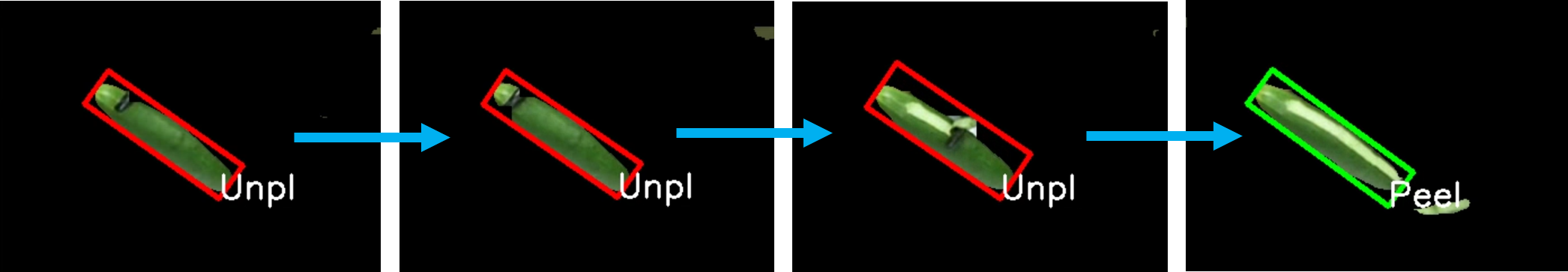}
        \caption{State of the object}
        \label{fig:sdemo}
    \end{subfigure}
    \caption{Progression of demonstration and corresponding states of the object. The red bounding boxes show that the state is unpeeled and the green bounding box indicates a peeled state.}
\end{figure}
We are more biased towards the first exploratory procedure and give it  more weightage in comparison with the second procedure.  An epsilon greedy approach is used to decide which procedure the RL agent performs. We are more biased towards exploring by adding a new action to our action set. This is because the RL agent at the very beginning (before going into exploratory procedures ) tried to exploit the current best option available in the q table and if that fails we assume that the other actions in the given set could also lead to failure as the exploited action was the most optimal action to take in comparison to all the other actions in the action set based on the learned policy at that point in time. Thus, the way we employ exploration and exploitation is to ensure that the robot doesn’t consume time mapping out the q values for the entire action space but rather focuses on the action space information it has currently in its possession. The skill and task policies are learned over time and exploration is performed only when required or only when the current best policy doesn’t result in task completion. 

\section{Experiment}
Our goal is to evaluate the effectiveness of the proposed one-shot learning and self-evaluation framework on a contact-intensive task. We looked at tasks involved in deploying a social robot in domestic settings. One common contact-intensive task that occurs in such environments is the task of peeling vegetables. Here, we consider peeling a cucumber in particular and specify how the learnt policies can be extended to vegetables of different stiffness.
\subsection{Demonstration}
A single demonstration of the task is performed by the agent. Importance is given to the task of peeling rather than peeling the entire vegetable through repeated actions, i.e. the agent performs only one peeling action which results in a partially peeled cucumber. The example of the demonstration used is shown in Fig. \ref{fig:demo}

\subsection{Inference} In the case of peeling task the policy learnt is $$ \Pi = \{(C_1, s*_1, \Theta_1), (C_2, s*_2, \Theta_2), (C_3, s*_3, \Theta_3)\}$$, where skill $C_1$ is \textit{approach} with object cucumber edge as reference, skill $C_2$ is \textit{move to contact}, and $C_3$ is \textit{contact trajectory} i.e. execute a trajectory learnt from demonstration with contact ensured.
The states of the object, in this case the cucumber is detected using features in the object property database. There are two possible states unpeeled and peeled. The states are inferred from vision and it is observed that there is no state change in cucumber after $C_1$ or $C_2$, and at end of $C_3$ the state changes from peeled to unpeeled. The change of state detected as the demonstration progresses is shown in Fig. \ref{fig:sdemo}. The state \textit{peel} is detected when atleast $10\%$ of the visible surface area is peeled.
\begin{figure}[t]
    \centering
    \includegraphics[width=0.5\textwidth]{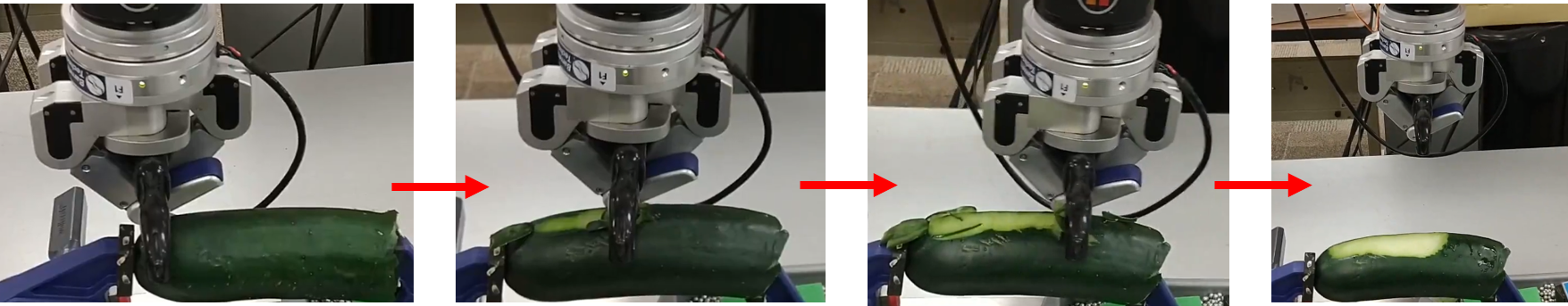}
    \caption{Execution of the task by robot agent, the figure shows the progression of skill over mulitple learning actions}
    \label{fig:act}
\end{figure}
\subsection{Execution} 
The execution of the learnt policy is done based on the control policy described in section \ref{skills}. The control policy for approach is very straight forward, it reduces to a PD control based on visual error. The controller for $move to contact$ uses normal force as the feedback condition, i.e. the sensor feedback function for this skill is given by, $f(x^d,s*) = F_n - F_d$ where $F_n$ is the normal reaction force felt at the end effector and $F_d$ is the desired/minimum contact force (0.5N in our case). The desired pose $x_d$ is relative position of hand w.r.t cucumber which is one edge of the cucumber.\\
For the \textit{move with contact} is a skill with impedance control model. The feedback error function in \ref{eqn:fski} is given by $f(x^d,s*) = \theta(x_t) - \theta(x_d)$, where $\theta(x)$ is the joint angles corresponding to the pose $x$ of the arm. The positional error to next point on the trajectory is computed using the internal pose estimates and the feedback error function $f$ is computed from this error and the error in applied contact force. The execution is shown in Fig. \ref{fig:act}

\subsection{Self Evaluation}
\begin{figure}[t]
    \centering
    \begin{subfigure}[t]{0.5\textwidth}
        \centering
        \includegraphics[width=0.5\textwidth]{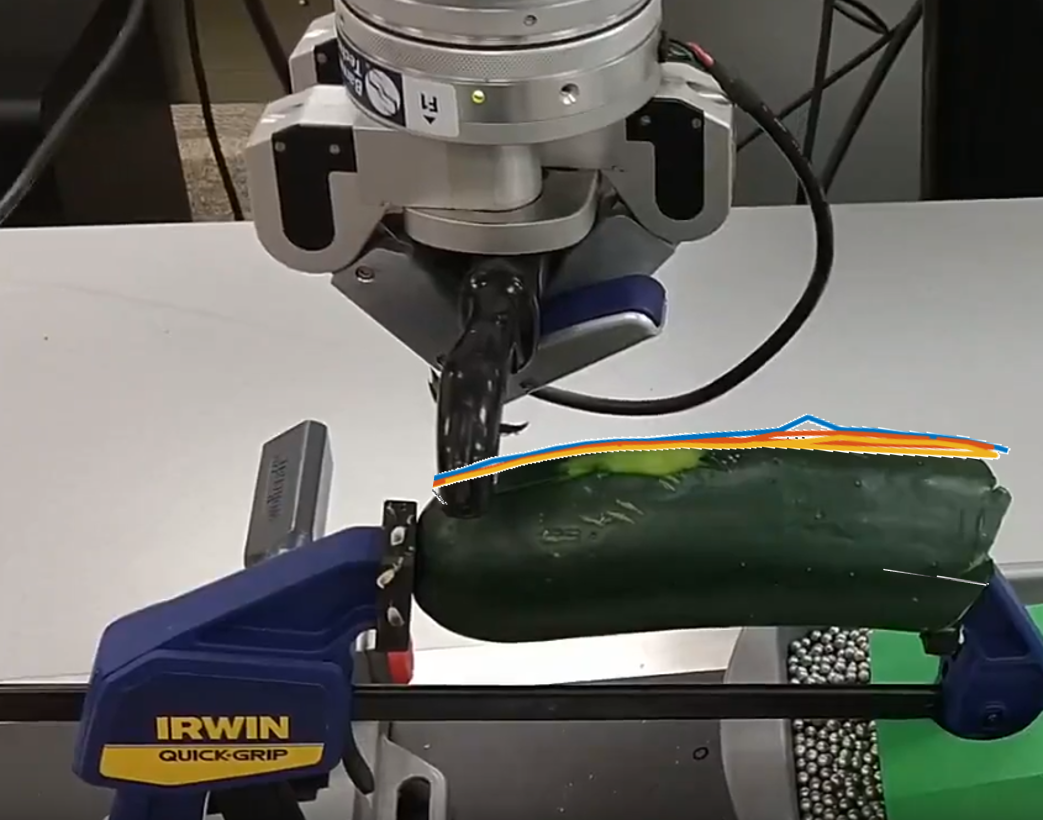}
        \caption{Action trajectories in the robot scene}
        \label{fig:r_act1}
    \end{subfigure}\\
    ~ 
    \begin{subfigure}[h]{0.5\textwidth}
        \centering
        \includegraphics[width=0.8\textwidth]{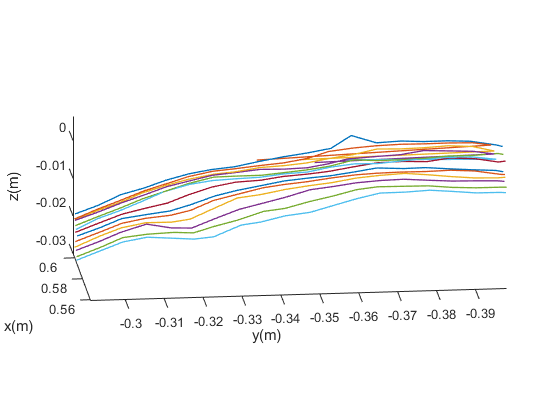}
        \caption{Set of explored contact trajectory actions}
        \label{fig:r_traj}
    \end{subfigure}
    \caption{The action set explored by the robot agent }
    \label{fig:r_act}
\end{figure}

Rewards are assigned depending on the completion of the skill i.e. the occurrence of a state transition from an unpeeled state to a peeled state. A positive reward of 2 is assigned when the robot peels the vegetable successfully and a negative reward of -5 is awarded in case of a failed execution. A discount factor of 0.3 was used so the robot agent focuses on achieving higher rewards in the short term as we are greedy towards skill completion. After the demonstration a baseline action trajectory is inferred which is then used to obtain the first contact trajectory which constitutes the first action in the action set. This contact trajectory is obtained with a constraint such that it achieves a normal contact force of 0.5N at every point along the trajectory. Every subsequent action added to the action set after the first action would aim at achieving a contact force which is 0.3N greater than its previous action (the last action in the action set) as seen in Fig. \ref{fig:r_act}. We deal with a noise measure of $\pm$0.2N  and hence we chose a contact force increment of 0.3N to represent the subsequent actions added to the action set.
In the case of the peeling task we consider the vegetable stiffness to be an important property. During task performance we do not want to apply too much force on a vegetable whose compliance is very high and risk damaging the vegetable. Understanding object properties helps us group vegetables of similar stiffness and treat these groups as a single entity. \cite{agrawal1997mechanical} \cite{williams2005mechanical} shows us how stiffness varies over vegetables. Using clustering algorithm we deduced two clusters of vegetables separated by stiffness. Our self evaluation framework would consider a separate Q table for each cluster.   \\

Exploration and Exploitation methods explained in section \ref{exp} are employed during the self evaluation process. In our case of a peeling task we start by exploiting the already learned policy and failure to complete the task with this policy results in two exploratory methods. The first method deals with adding a new action to the action set and exploring that action. The robot would have to add a new action to the action set which would be the contact trajectory achieved at a higher force than all existing actions. Intuitively we know that this would be the best way to explore (in comparison to the second exploratory method that deals with randomly exploring one of the already explored action again). \\

\section{Conclusions}

This paper presents a one-shot learning from demonstration framework to learn contact-intensive tasks and employs a self evaluation routine to optimize the contact-intensive skills corresponding to the task. The proposed method was implemented in the task of peeling a vegetable. The vegetable was peeled successfully from the inference obtained during demonstration and self evaluation successfully tuned the skill parameters with respect to completion of all skill-goals. By using a 'greedy-towards-skill-goal' approach, during self evaluation the robot explores and exploits the environment such that it focuses on the immediate completion of the skill. This tuning can be aligned with skill performance rather than just skill completion by introducing feedback in evaluation through the process of coaching, resulting in more intuitive task programming paradigms.
\bibliographystyle{IEEEtran}
\bibliography{refs}

\begin{thebibliography}{10}
\providecommand{\url}[1]{#1}
\csname url@rmstyle\endcsname
\providecommand{\newblock}{\relax}
\providecommand{\bibinfo}[2]{#2}
\providecommand\BIBentrySTDinterwordspacing{\spaceskip=0pt\relax}
\providecommand\BIBentryALTinterwordstretchfactor{4}
\providecommand\BIBentryALTinterwordspacing{\spaceskip=\fontdimen2\font plus
\BIBentryALTinterwordstretchfactor\fontdimen3\font minus
  \fontdimen4\font\relax}
\providecommand\BIBforeignlanguage[2]{{%
\expandafter\ifx\csname l@#1\endcsname\relax
\typeout{** WARNING: IEEEtran.bst: No hyphenation pattern has been}%
\typeout{** loaded for the language `#1'. Using the pattern for}%
\typeout{** the default language instead.}%
\else
\language=\csname l@#1\endcsname
\fi
#2}}

\bibitem{cakmak2010exploiting}
M.~Cakmak, N.~DePalma, R.~I. Arriaga, and A.~L. Thomaz, ``Exploiting social
  partners in robot learning,'' \emph{Autonomous Robots}, vol.~29, no. 3-4, pp.
  309--329, 2010.

\bibitem{dufay1984approach}
B.~Dufay and J.-C. Latombe, ``An approach to automatic robot programming based
  on inductive learning,'' \emph{The International journal of robotics
  research}, vol.~3, no.~4, pp. 3--20, 1984.

\bibitem{muench1994robot}
S.~Muench, J.~Kreuziger, M.~Kaiser, and R.~Dillman, ``Robot programming by
  demonstration (rpd)-using machine learning and user interaction methods for
  the development of easy and comfortable robot programming systems,'' in
  \emph{Proceedings of the International Symposium on Industrial Robots},
  vol.~25.\hskip 1em plus 0.5em minus 0.4em\relax International Federation of
  Robotics, \& Robotic Industries, 1994, pp. 685--685.

\bibitem{argall2011teacher}
B.~D. Argall, B.~Browning, and M.~M. Veloso, ``Teacher feedback to scaffold and
  refine demonstrated motion primitives on a mobile robot,'' \emph{Robotics and
  Autonomous Systems}, vol.~59, no. 3-4, pp. 243--255, 2011.

\bibitem{derimis2002imitations}
Y.~Derimis and G.~Hayes, ``Imitations as a dual-route process featuring
  predictive and learning components: a biologically plausible computational
  model,'' \emph{Imitation in animals and artifacts}, pp. 327--361, 2002.

\bibitem{abbeel2004apprenticeship}
P.~Abbeel and A.~Y. Ng, ``Apprenticeship learning via inverse reinforcement
  learning,'' in \emph{Proceedings of the twenty-first international conference
  on Machine learning}.\hskip 1em plus 0.5em minus 0.4em\relax ACM, 2004, p.~1.

\bibitem{billard2016learning}
A.~G. Billard, S.~Calinon, and R.~Dillmann, ``Learning from humans,'' in
  \emph{Springer handbook of robotics}.\hskip 1em plus 0.5em minus 0.4em\relax
  Springer, 2016, pp. 1995--2014.

\bibitem{stulp2012reinforcement}
F.~Stulp, E.~A. Theodorou, and S.~Schaal, ``Reinforcement learning with
  sequences of motion primitives for robust manipulation,'' \emph{IEEE
  Transactions on robotics}, vol.~28, no.~6, pp. 1360--1370, 2012.

\bibitem{voyles1999gesture}
R.~M. Voyles and P.~K. Khosla, ``Gesture-based programming: A preliminary
  demonstration,'' in \emph{Proceedings 1999 IEEE International Conference on
  Robotics and Automation (Cat. No. 99CH36288C)}, vol.~1.\hskip 1em plus 0.5em
  minus 0.4em\relax IEEE, 1999, pp. 708--713.

\bibitem{elliott2017learning}
S.~Elliott, Z.~Xu, and M.~Cakmak, ``Learning generalizable surface cleaning
  actions from demonstration,'' in \emph{2017 26th IEEE International Symposium
  on Robot and Human Interactive Communication (RO-MAN)}.\hskip 1em plus 0.5em
  minus 0.4em\relax IEEE, 2017, pp. 993--999.

\bibitem{levine2015learning}
S.~Levine, N.~Wagener, and P.~Abbeel, ``Learning contact-rich manipulation
  skills with guided policy search,'' in \emph{2015 IEEE international
  conference on robotics and automation (ICRA)}.\hskip 1em plus 0.5em minus
  0.4em\relax IEEE, 2015, pp. 156--163.

\bibitem{baisero2015temporal}
A.~Baisero, Y.~Mollard, M.~Lopes, M.~Toussaint, and I.~L{\"u}tkebohle,
  ``Temporal segmentation of pair-wise interaction phases in sequential
  manipulation demonstrations,'' in \emph{2015 IEEE/RSJ International
  Conference on Intelligent Robots and Systems (IROS)}.\hskip 1em plus 0.5em
  minus 0.4em\relax IEEE, 2015, pp. 478--484.

\bibitem{elliott2017efficient}
S.~Elliott, R.~Toris, and M.~Cakmak, ``Efficient programming of manipulation
  tasks by demonstration and adaptation,'' in \emph{2017 26th IEEE
  International Symposium on Robot and Human Interactive Communication
  (RO-MAN)}.\hskip 1em plus 0.5em minus 0.4em\relax IEEE, 2017, pp. 1146--1153.

\bibitem{mollard2015robot}
Y.~Mollard, T.~Munzer, A.~Baisero, M.~Toussaint, and M.~Lopes, ``Robot
  programming from demonstration, feedback and transfer,'' in \emph{2015
  IEEE/RSJ International Conference on Intelligent Robots and Systems
  (IROS)}.\hskip 1em plus 0.5em minus 0.4em\relax IEEE, 2015, pp. 1825--1831.

\bibitem{cabrera2017one}
M.~E. Cabrera, N.~Sanchez-Tamayo, R.~Voyles, and J.~P. Wachs, ``One-shot
  gesture recognition: One step towards adaptive learning,'' in \emph{2017 12th
  IEEE International Conference on Automatic Face \& Gesture Recognition (FG
  2017)}.\hskip 1em plus 0.5em minus 0.4em\relax IEEE, 2017, pp. 784--789.

\bibitem{wan2016explore}
J.~Wan, G.~Guo, and S.~Z. Li, ``Explore efficient local features from rgb-d
  data for one-shot learning gesture recognition,'' \emph{IEEE transactions on
  pattern analysis and machine intelligence}, vol.~38, no.~8, pp. 1626--1639,
  2016.

\bibitem{lee2015learning}
A.~X. Lee, H.~Lu, A.~Gupta, S.~Levine, and P.~Abbeel, ``Learning force-based
  manipulation of deformable objects from multiple demonstrations,'' in
  \emph{2015 IEEE International Conference on Robotics and Automation
  (ICRA)}.\hskip 1em plus 0.5em minus 0.4em\relax IEEE, 2015, pp. 177--184.

\bibitem{taa2018wm}
T.~Soratana, M.~V. S.~M. Balakuntala, P.~Abbaraju, R.~Voyles, J.~Wachs, and
  M.~Mahoor, ``Glovebox handling of high-consequence materials with super
  baxter and gesture-based programming,'' in \emph{44th Annual Waste Management
  Symposium (WM2018)}, 2018.

\bibitem{Watkins1992}
\BIBentryALTinterwordspacing
C.~J. C.~H. Watkins and P.~Dayan, ``Q-learning,'' \emph{Machine Learning},
  vol.~8, no.~3, pp. 279--292, May 1992. [Online]. Available:
  \url{https://doi.org/10.1007/BF00992698}
\BIBentrySTDinterwordspacing

\bibitem{agrawal1997mechanical}
K.~Agrawal, P.~Lucas, J.~Prinz, and I.~Bruce, ``Mechanical properties of foods
  responsible for resisting food breakdown in the human mouth,'' \emph{Archives
  of Oral Biology}, vol.~42, no.~1, pp. 1--9, 1997.

\bibitem{williams2005mechanical}
S.~H. Williams, B.~W. Wright, V.~d. Truong, C.~R. Daubert, and C.~J. Vinyard,
  ``Mechanical properties of foods used in experimental studies of primate
  masticatory function,'' \emph{American Journal of Primatology: Official
  Journal of the American Society of Primatologists}, vol.~67, no.~3, pp.
  329--346, 2005.

\end{thebibliography}
\end{document}